\documentclass[conference]{IEEEtran}
\IEEEoverridecommandlockouts
\usepackage{cite}
\usepackage{subfigure}
\usepackage{lscape}
\usepackage{amsmath,amssymb,amsfonts}
\usepackage{graphicx}
\usepackage{pstricks-add}
\usepackage{textcomp}
\usepackage{algorithm}
\usepackage{multirow}
\usepackage{tabularx}
\usepackage{graphicx}
\usepackage{enumitem}
\usepackage{algpseudocode}
\usepackage{xcolor}
\def\BibTeX{{\rm B\kern-.05em{\sc i\kern-.025em b}\kern-.08em
    T\kern-.1667em\lower.7ex\hbox{E}\kern-.125emX}}

\begin{document}

\title{Centralized vs Decentralized Federated Learning: A trade-off performance analysis\\
}

\author{\IEEEauthorblockN{ Chaimaa MEDJADJI}
\IEEEauthorblockA{\textit{University of Luxembourg} \\
chaimaa.medjadji@uni.lu}
\and
\IEEEauthorblockN{Guilain LEDUC}
\IEEEauthorblockA{\textit{University of Luxembourg} \\
guilain.leduc@uni.lu}
\and
\IEEEauthorblockN{Sylvain KUBLER}
\IEEEauthorblockA{\textit{University of Luxembourg} \\
sylvain.kubler@uni.lu}
\and
\IEEEauthorblockN{Yves Le Traon}
\IEEEauthorblockA{\textit{University of Luxembourg} \\
Yves.LeTraon@uni.lu}
}

\maketitle

\begin{abstract}

Federated Learning (FL) has emerged as a promising paradigm for collaborative model training across distributed edge devices while preserving data privacy especially with the huge increase amount of data due to the adoption of technologies which contributes to the growing number of IoT devices. Storing this amount of data centrally is challenging due to issues like limited communication, privacy, and regulations. FL can be Centralized (CFL), Decentralized (DFL), and Semi-decentralized (SDFL). Choosing the right FL architecture depends on the application's needs. However, very few research studies have experimentally compared these three types of architectures to not only understand the respective strengths and limitations, but also trade-offs between different performance indicators. This paper overcome this lack of analysis, conducting experimental analyses using the Fedstellar simulator, MNIST dataset, and MLP classifier.
\end{abstract}

\vspace{1cm}
\section{Introduction}
The rise of emerging application scenarios, such as augmented and virtual reality, autonomous driving, and Digital Twin, has triggered a substantial surge in the number of Internet of Things (IoT) devices.
The International Data Corporation (IDC) estimates that 55.7 billion IoT devices will be connected by 2025, generating almost 80B zettabytes (ZB) of data.
This data explosion will contribute to a total global data storage exceeding 175 ZB. 

Artificial Intelligence (AI), particularly Machine Learning (ML), is well-positioned to the immense volumes of data anticipated in the coming years. However, data from a multitude of IoT devices are often stored in distributed architectures for diverse scenarios such as smart grids, remote health monitoring, or the Internet of Vehicles. Challenges such as limited communication resources, data privacy concerns, and country-specific regulations render the centralized collection of data impractical or inefficient, as traditionally practiced in ML. In 2016, Google emerged Federated Learning (FL) \cite{kon16} as a solution to this challenge, enabling clients -- \emph{referred to as participants or nodes of a federation} -- to collaboratively train models without the need to share raw training data. FL can be categorized into three types of architectures, namely: Centralized (CFL), Decentralized (DFL), and a combinaison of the two known as Semi-decentralized (SDFL).

The choice of using one type of architecture or another is highly dependent on the application requirements and constraints. However, there are other aspects to take into consideration as well, such as how a given FL architecture impacts on the application performance. A recent survey \cite{bel23} has emphasized the intricate nature of DFL and the multitude of Key Performance Indicators (KPIs) -- \emph{over 40} -- to be taken into account when evaluating (D)FL platforms. While research studies have focused on the architectural design of CFL, DFL and SDFL, very little research, if any, has focused on experimentally comparing these architectures. Understanding how CFL, DFL, and SDFL perform across various KPIs is crucial, as it can reveal potential trade-offs or synergies among them. This paper presents the initial experimental analyses in this context.

Section~\ref{Sec:Biblio} provides the necessary background about CFL, DFL and SDFL. Section~\ref{Sec:Methodo} gives insight into the methodology underlying our experimental analyses. Section~\ref{Sec:Results} presents the results of our experiments conducted with the Fedstellar simulator, MNIST dataset, and MLP classifier MNIST data. The conclusion is provided in section~\ref{Sec:conclu}.




\section{Related works}\label{Sec:Biblio}
Section~\ref{Sec:Biblio:FL} provides a comprehensive overview of various FL architectures, including CFL, DFL, and SDFL. In Section~\ref{Sec:Biblio:CFLvsDFL}, the discussion delves into the expanding body of literature regarding DFL and SDFL, along with investigations that compare these architectures with CFL.

\begin{figure*}[t]
 \centering
 \subfigure[\label{fig:FL:CFL} Centralized FL ]{
 \psset{xunit=1mm,yunit=1mm,runit=1mm}
 \scalebox{1.45}{
\begin{pspicture}(0,0)(35,27)
 \scriptsize
 \put(-5,0){\includegraphics[clip,trim=0mm 0mm 5mm 0mm, height=30mm]{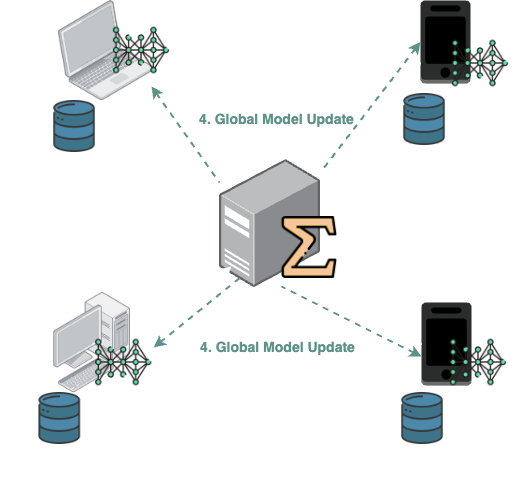}}
 \end{pspicture}
 }}
 \subfigure[\label{fig:FL:DFL} Decentralized FL ]{
 \psset{xunit=1mm,yunit=1mm,runit=1mm}
 \scalebox{1.45}{
\begin{pspicture}(0,0)(45,27)
 \scriptsize
 \put(-5,0){\includegraphics[clip,trim=2mm 0mm 0mm 0mm, height=30mm]{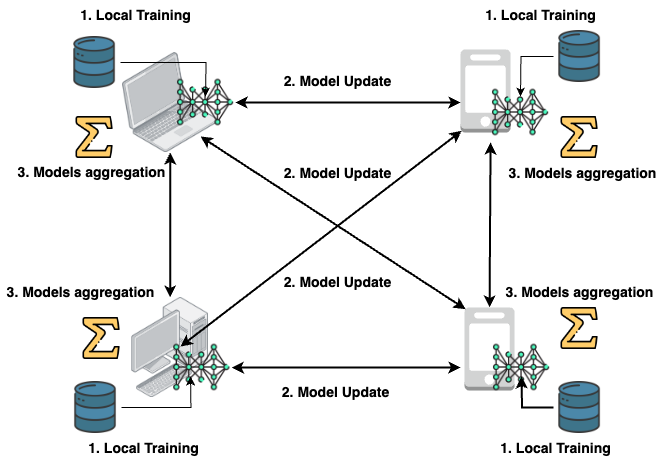}}
 \end{pspicture}
 }}
 \subfigure[\label{fig:FL:SDFL}Semi-Decentralized FL ]{
 \psset{xunit=1mm,yunit=1mm,runit=1mm}
 \scalebox{1.45}{
\begin{pspicture}(0,0)(30,27)
 \scriptsize
 \put(-5,0){\includegraphics[clip,trim=2mm 0mm 0mm 0mm, height=30mm]{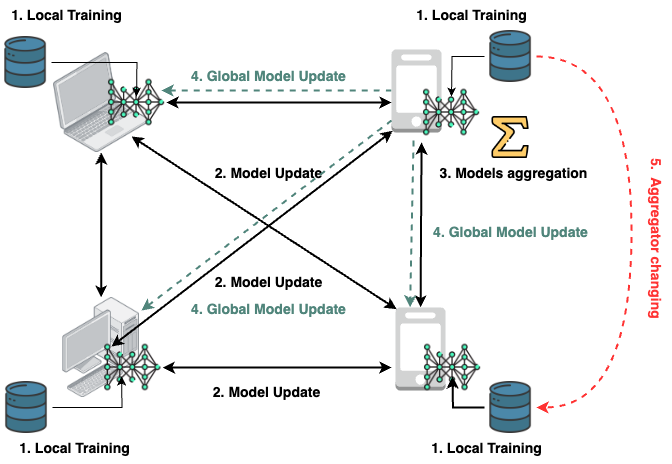}}
 \end{pspicture}
 }}
 \caption{Federated Learning Architectures}\label{fig:FL}
 \end{figure*}

\subsection{Federated Learning (FL) architecture types}\label{Sec:Biblio:FL}
FL is a ML setting to synthesize global models from local models trained on the edge. FL was first developed by Google in 2016 \cite{kon16} for their Gboard application, to learn new words and phrases. FL is a decentralized training approach where an iterative process computes model training updates at the edge, aggregating these updates to produce the global update to be applied to the model. This core concept of aggregating model updates was key in allowing for a single global model to be produced from edge training. To build this global model, an aggregation mechanism is implemented to merge in a consistent manner the set of local models produced at some iteration. The most well-known aggregation algorithm, popular for its simplicity and efficiency, is called federated averaging (FedAvg). FedAvg computes the parameter-wise arithmetic mean across the local models.

Currently, most of the FL applications rely on a CFL architecture, as illustrated in \figurename~\ref{fig:FL:CFL}.
CFL has gained a significant attention from academia and industry \cite{tli20}. Lim et al. \cite{lim20} mentioned that researchers mainly focused on CFL applications in mobile networks across various end-user scenarios : the application of FL in broader contexts involving heterogeneous IoT devices (\cite{ngu21}), security and privacy concerns (\cite{khan21}), proposition of secure protocols between servers and participants (\cite{moth21}). Furthermore, other studies, such as \cite{josh22}, explored the applicability of CFL in specific medical scenarios, emphasizing privacy-preserving mechanisms for sensitive data. Witt et al. \cite{witt22} detailed commonly used frameworks in CFL applications. 

CFL offers certain conveniences, but also comes with its own set of disadvantages. Li et al. \cite{Li20} highlight the main challenges faced by CFL caused mainly by the server as third trust party and single point of failure. These challenges include the increase of data transmissions, especially when dealing with multiple participating nodes. Li et al. \cite{Li20} also point out the problem of reliance on a central server that can introduce communication bottlenecks and increased network traffic when faced with a large number of participants, leading to higher latency and reduced efficiency \cite{zh19}. As another problem of CFL, the security of the application can be impacted if the central server is maliciously compromised \cite{ka21}. To overcome the above-mentioned limitations, DFL was introduced in 2018 by Lalitha et al. \cite{la18}. DFL is based on the decentralized aggregation of model parameters among neighboring participants, eliminating the necessity for a central server, as depicted in \figurename~\ref{fig:FL:DFL}. It ensures collaborative learning by leveraging local computation and exchanging model parameters between nodes. Its operation revolves around swiftly transmitting updates computed locally by each node to the broader network. These local transmissions enable adjacent nodes to enhance their models, culminating in the creation of a more intelligent and privacy-preserving collaborative model.

In contrast to the previous approaches of FL, the SDFL architecture \cite{roy19} is a hybrid of both architectures CFL and DFL, as illustrated in \figurename~\ref{fig:FL:SDFL}. It maintain the aggregator role which rotates randomly among participants at each round of the learning process.




\subsection{CFL vs. DFL: current state of affairs}\label{Sec:Biblio:CFLvsDFL}
In contrast to CFL, DFL and SDFL were first employed to prevent communication bottleneck situations, which can quickly occur in CFL and result in performance issues \cite{yem22}.
As of today, most research on DFL and SDFL focuses on analyzing the usage of Distributed Ledger Technology (DLT) such as Blockchain \cite{you20, you22}. Also, authors have explored the applicability of DFL in specific domains such as the Internet of Vehicles (IoV) \cite{bill22}, wireless communications \cite{gupta22}, and Unmanned Aerial Vehicle (UAV) devices \cite{sara22}. 

These studies investigate the performance of DFL in targeted application scenarios, but rarely, not to say never,  experimentally compare the pros and cons of DFL over CFL, and vice-versa (e.g., identification of trade-offs between different performance indicators, aka KPIs). Although decentralized methods showing promise in achieving comparable convergence to centralized methods with reduced communication, their test performance remains inefficient in empirical studies.

\begin{figure*}[ht]
\centering
\includegraphics[width=1\textwidth]{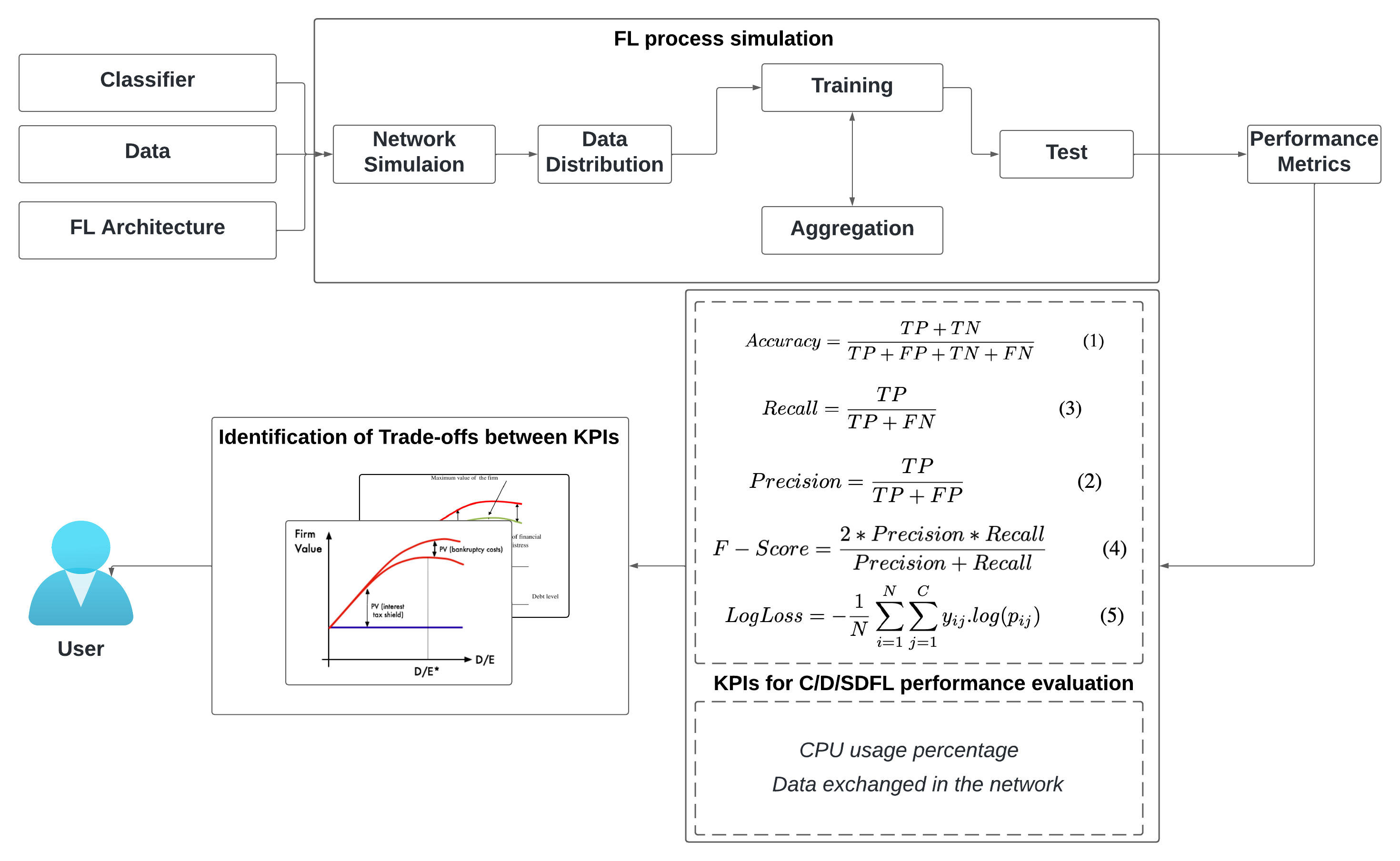}
\caption{Methodology underlying the method of evaluating and identifing the trade-offs between different KPIs.}
\label{fig:Methodo} 
\end{figure*}

\section{Performance comparison methodology}\label{Sec:Methodo}

To evaluate to what extent the different FL architectures (CFL, DFL, SDFL) perform well under specific application scenarios, and analyze/identify possible trade-offs in performance, the methodology depicted in \figurename~\ref{fig:Methodo} is defined. The first stage consists in simulating the FL process. The second stage involves implementing relevant KPIs for CFL, DFL and SDFL performance evaluation. The third stage consists in identifying the trade-offs that may possibly appear between the implemented KPIs. The three stages are elaborated on in sections~\ref{Sec:Method:Stage1} to \ref{Sec:Method:Stage3}. As of today, to the best of our knowledge, only the FedStellar platform \cite{bel24} supports the simulation of all three FL architectures, enabling the testing of highly scalable network configurations. Additionally, it facilitates the monitoring of multiple KPIs throughout various stages of the FL process. This leads us to choose FedStellar as the foundational simulation layer for our comparative analysis methodology.

\subsection{Stage 1: FL process simulation}\label{Sec:Method:Stage1}
Due to the intricate nature of implementing the three types of FL architectures, scholars have initiated the development of simulators for evaluation purposes. Several FL simulator platforms have been introduced in recent years, as outlined in \tablename~\ref{Tab:Platforms}. This table highlights the characteristics of each platform, including: \emph{(i) FL architectures:} the supported types of architecture (CFL and/or DFL and/or SDFL); \emph{(ii) Scalability:} the platform's ability to accommodate large FL networks (classified on a 3-scale basis: low, medium, high); \emph{(iii) Metrics:} which can be employed within the solution to assess the FL process; \emph{(iv) Open source:} indicating whether the simulator is open source or proprietary.
\begin{table*}[!ht]
\centering
\caption{Comparison of some FL frameworks.}\label{Tab:Platforms}
\begin{tabular}{llclc}
\textbf{Solution}                              & \textbf{FL architecture} & \textbf{Scalability} & \textbf{Evaluation Metrics}                       & \textbf{Open source} \\ 
\hline
{\textbf{TFF}}            & CFL                      & Medium              & Model performance, Convergence Time               & Yes                  \\ 
{\textbf{FATE}}           & CFL                      & Medium               & Model performance                                 & Yes                  \\ 
{\textbf{BrainTorrent}}   & SDFL                     & -        & Model performance, CPU usage                      & No                   \\
{\textbf{ScatterBrained}} & DFL                      & Low                  & Model performance                                 & Yes                  \\ 
{\textbf{PySyft}}         & CFL, DFL                 & Medium               & Model performance                                 & Yes                  \\ 
{\textbf{IPLS}}           & DFL                      & Medium               & Model performance (Accuracy)                      & Yes                  \\ 
{\textbf{2DF-IDS}}        & DFL                      & Medium               & Model performance                                 & No                   \\ 
{\textbf{P2PFL}}          & DFL                      & Medium               & Model performance (Accuracy, Loss)                & Yes                  \\ 
{\textbf{FedStellar}}     & CFL, DFL, SDFL           & High                 & Model performance, Resources usage, Communication & Yes                  \\ \hline
\end{tabular}
\end{table*}
\\

\begin{figure*}[t]
\centering
\includegraphics[width=0.75\textwidth]{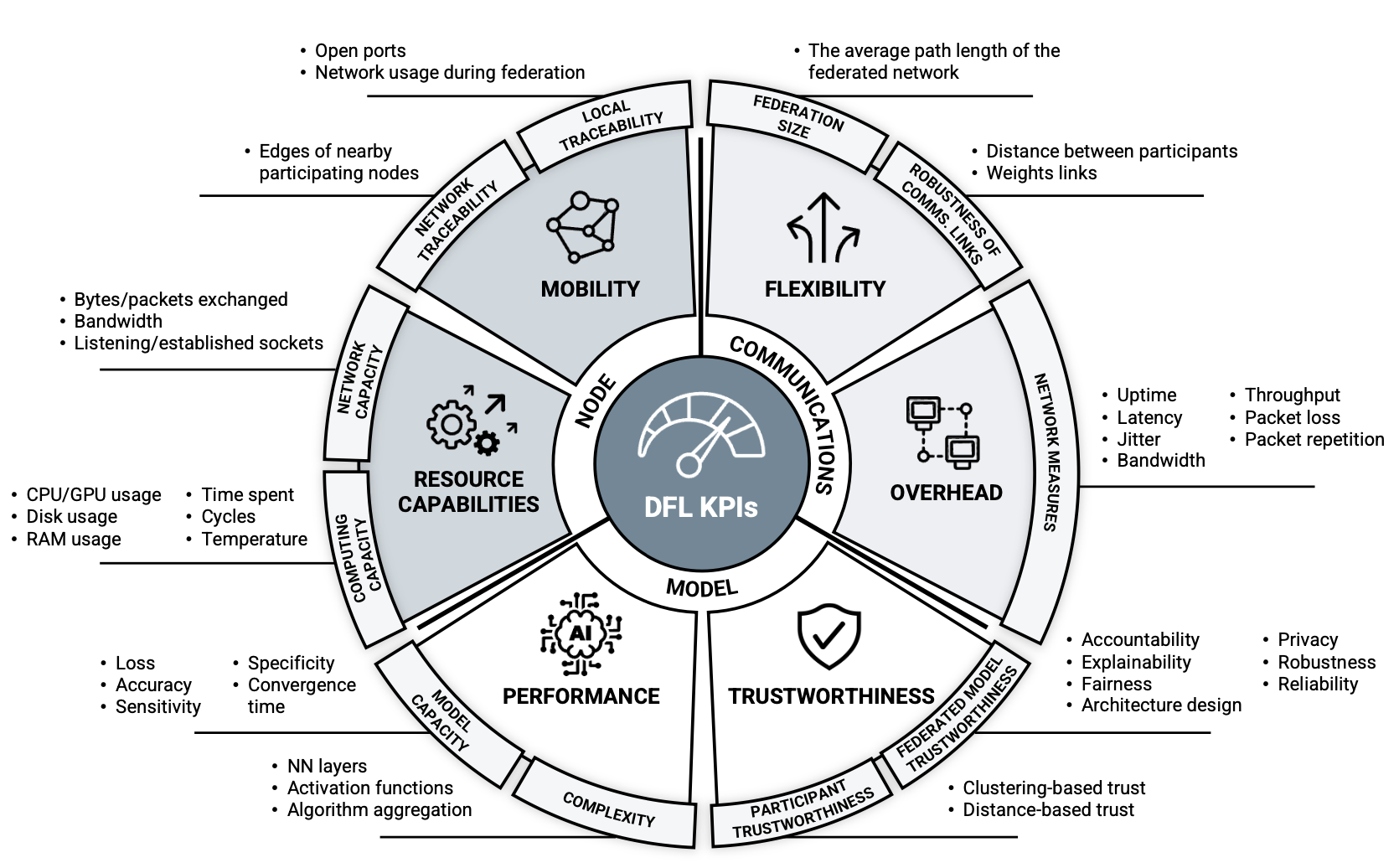}
\caption{KPIs introduced by Beltran et al. \cite{bel23} to evaluate (D)FL architectures/implementations.}\label{fig:rosece} 
\end{figure*}

\subsection{Stage 2: KPIs for CFL, DFL, and SDFL performance evaluation} \label{Sec:Method:Stage2}
The recent survey of Beltran et al. \cite{bel23} has proposed a taxonomy of KPIs, along with categories, to cover the different aspects of FL, as shown in \figurename~\ref{fig:rosece}. KPIs are categorized in three categories:
\begin{itemize}
    \item \textbf{Node:} it evaluates the performance of participating nodes within the FL network. The heterogeneity and dynamism of the participating nodes introduces complexity to the network, making it crucial to identify relevant factors for evaluation. The Node category is declined into: (i) \textit{Resource capabilities:} evaluate the computational power and the network capacity; \textit{(ii) Node mobility:} evaluate the high dynamism of the network.
    \item \textbf{Communications:} it evaluates the extent to which the communication layer is efficient and effective. Indeed, the exchange of model parameters between nodes can lead to instability that may compromise the reliability of the system/application. The Communications category is declined into: \textit{(i) Communication flexibility:} evaluate the FL network capacity to adapt to dynamic network conditions and maintain stable communication channels; \textit{(ii) Network overhead:} quantify the additional resources consumed by communication processes beyond the essential data exchange. Evaluating this overhead helps in measuring the efficiency and scalability of the FL network.
    \item \textbf{Model:} it evaluates the efficiency and effectiveness of the FL model to accomplish its learning task. It divides into \textit{(i) Performance:} evaluate the efficiency of the FL model; \textit{(ii) Trustworthiness:} evaluate trustworthiness aspects of AI systems, including transparency, robustness, fairness, and so forth \cite{Da22}.
\end{itemize}

In total, over $40$~KPIs have been delineated in \cite{bel23} across the aforementioned three categories, recognizing that some KPIs may encompass several others (e.g., explainability could be assessed through faithfulness, stability, transferability, or compactness of the explanation). It is important to note that the three categories identified by \cite{bel23} may be perceived as different in nature: Node and Communications KPIs may be viewed as constrained by the underlying or implemented network infrastructure, while Model KPIs may be seen as the ultimate objective of the application (ensuring adherence to the FL model requirements).

While our study's long-term objective is to encompass all categories and their associated KPIs, we are currently constrained -- due to the simulator -- in the number of KPIs we can measure or assess at this stage. Presently, FedStellar only permits the experimental measurement of (1) accuracy, (2) precision, (3) recall, (4) F-score, (5) loss, (6) CPU usage percentage, and (7) bytes exchanged between participants. 

\subsection{Stage 3: Identification des trade-offs between the KPIs}\label{Sec:Method:Stage3}
The last stage consists in anlayzing possible trade-offs between KPIs, and highlight the key distinctions among the different FL architectures under specific application and network conditions.

\begin{figure*}[!ht]
 \centering
 \subfigure[\label{fig:fl3:accuracy} Accuracy]{
 \psset{xunit=1mm,yunit=1mm,runit=1mm}
 \scalebox{1.6}{
\begin{pspicture}(0,0)(33,33)
 \scriptsize
 \put(-5.5,0){\includegraphics[clip,trim=0mm 0mm 5mm 0mm, height=30mm]{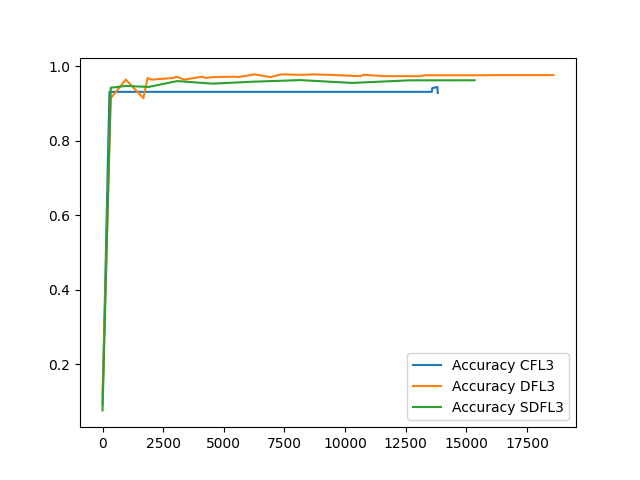}}
 \end{pspicture}
 }}\hfill
 \subfigure[\label{fig:fl3:loss} Loss]{
 \psset{xunit=1mm,yunit=1mm,runit=1mm}
 \scalebox{1.6}{
\begin{pspicture}(0,0)(33,33)
 \scriptsize
 \put(-3.5,0){\includegraphics[clip,trim=2mm 0mm 0mm 0mm, height=30mm]{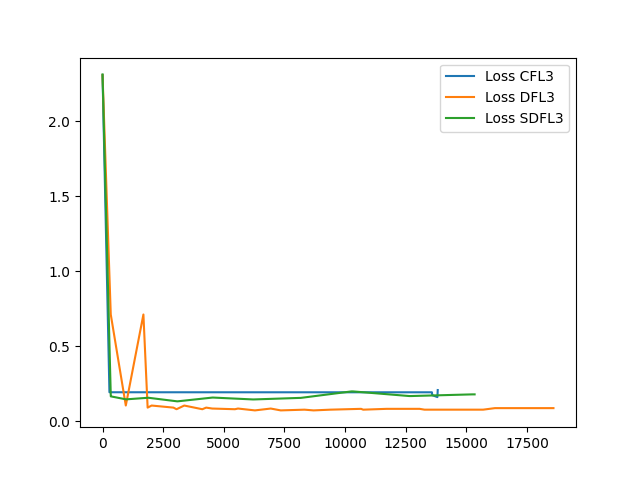}}
 \end{pspicture}
 }}\hfill
 \subfigure[\label{fig:fl3:precision} Precision]{
 \psset{xunit=1mm,yunit=1mm,runit=1mm}
 \scalebox{1.6}{
\begin{pspicture}(0,0)(33,30)
 \scriptsize
 \put(-1.5,0){\includegraphics[clip,trim=0mm 0mm 0mm 0mm, height=30mm]{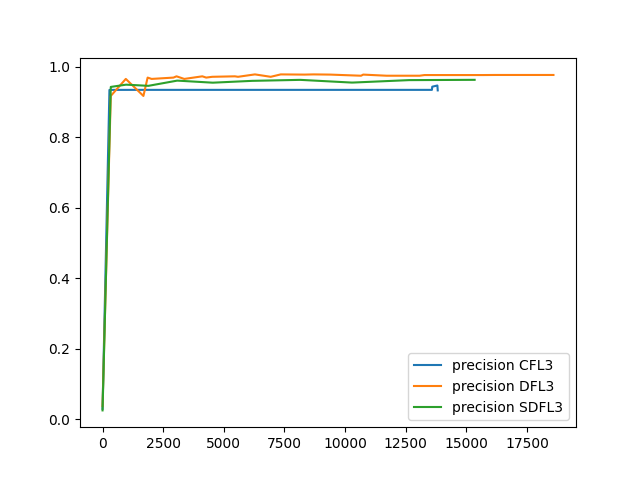}}
 \end{pspicture}
 }}\hfill
  \subfigure[\label{fig:fl3:recall} Recall]{
 \psset{xunit=1mm,yunit=1mm,runit=1mm}
 \scalebox{1.6}{
\begin{pspicture}(0,0)(33,26)
 \scriptsize
 \put(-5.5,0){\includegraphics[clip,trim=0mm 0mm 0mm 10mm, height=27mm]{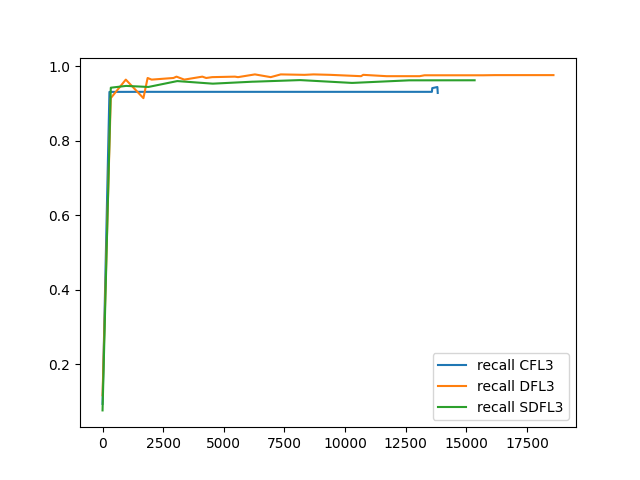}}
 \end{pspicture}
 }}\hfill
 \subfigure[\label{fig:fl3:bytes} Bytes exchanged in the FL network]{
 \psset{xunit=1mm,yunit=1mm,runit=1mm}
 \scalebox{1.6}{
\begin{pspicture}(0,0)(33,26)
 \scriptsize
 \put(-7.5,0){\includegraphics[clip,trim=0mm 0mm 0mm 10mm, height=26mm]{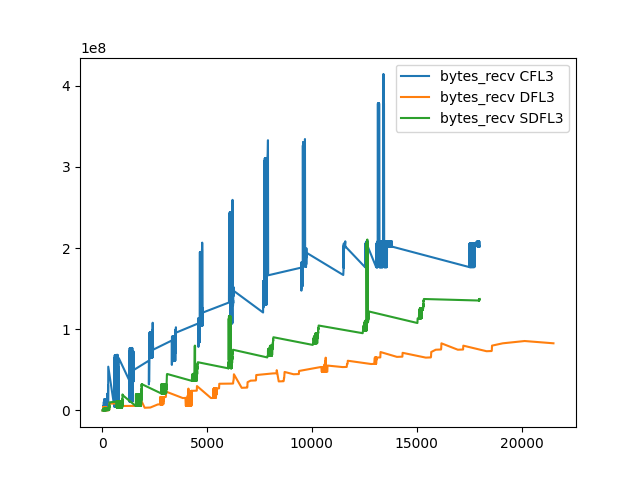}}
 \end{pspicture}
 }}\hfill
 \subfigure[\label{fig:fl3:cpu} CPU usage percentage]{
 \psset{xunit=1mm,yunit=1mm,runit=1mm}
 \scalebox{1.6}{
\begin{pspicture}(0,0)(25,26)
 \scriptsize
 \put(-7.5,0){\includegraphics[clip,trim=0mm 0mm 0mm 2mm, height=26mm]{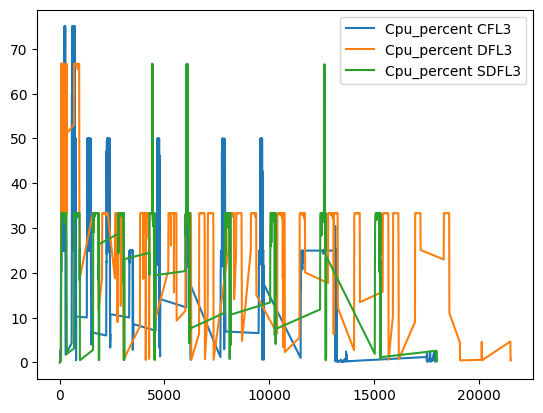}}
 \end{pspicture}
 }}
 \caption{Performance Evaluation Metrics with $|N|=3$ : (a) Accuracy  (b) Loss (c) Precision (d) Recall (e) Bytes exchanged in the FL network (f) CPU usage percentage.
 The DFL is the most efficient one with an accuracy of 97,84\%, a loss of 0.072925 using the less resources usage. The SDFL can reach a very close performance to the DFL one with an accuracy of 96.31\% and a loss of 0.133429, but it requires more time (see \tablename ~\ref{table:1}). CFL converges to the performance faster than the others and it is more stable but it is less efficient with an accuracy of 94.45\% and a loss of 0.161705, and it is the greediest one in terms of resources usage.
}\label{fig:fl3}
 \end{figure*}
 
\section{Experiments and results}\label{Sec:Results}
A set of experiments were conducted to evaluate the performance of the three types of FL architectures in a given use case scenario. Section~\ref{Sec:Results:setup} outlines the experimental setup. Section~\ref{Sec:Results:res}) delves into the results of the measured KPIs, with a focus on model performance, communication cost, and scalability.

\subsection{Experimental setup}\label{Sec:Results:setup}
Fedstellar simulator was used to implement a given use case scenario. In this regard, the MNIST dataset is used, which contains $60.000$~images of handwritten digits categorized into $10$~classes (0 to 9). In our experiments, the MNIST dataset is partitioned into subsets corresponding to the number of participants ($|N|$), with each participant ($n_{i}$) being allocated a subset comprising approximately $d_{i} = 60000/|N|$ samples, ensuring uniform distribution of labels. Multilayer Perceptron (MLP) \cite{pop09} is used as classification model in our experiments.

Throughout the experiments, we maintained a fixed number of rounds (10 rounds) and epochs (3 epochs). The network topology of participants in FL is configured as fully connected to ease communication and collaboration between participants.

To investigate the influence of scalability on the FL process, we alter the number of participants in each experiment. Beginning with $3$~participants due to constraints in computer capacity, we gradually increase to $4$, then to $6$, and finally to $8$ participants.

\begin{table}[t]
 \caption{Accuracy (\%) and Convergence time results. With the increase of the participants' number, the model's performance is declining and the convergence time is increasing for both CFL and SDFL due to the larger network size. In contrast, DFL demonstrates a decrease in convergence time  because of the smaller size of local data. This highlights DFL as a promising solution for scalability challenges. }

\label{table:1}
\begin{center}
 \begin{tabular}{ c  c  c  c } 
& Number of 
 Participants & Model (Accuracy \%) & Time (min) \\ [0.5ex] 
 \hline
 & 3 & $94.45\%$ & $\approx 18$ \\ 
  CFL & 4 & $91.47\%$ & $\approx 19$ \\ 
  & 6 & $90.78\%$ & $\approx 20$ \\ 
  & 8 & $89.83\%$ & $\approx 22$ \\ 
  \hline
  & 3 & $97.84\%$ & $\approx 43$ \\ 
  DFL & 4 & $97.61\%$ & $\approx 41$ \\ 
  & 6 & $97.5\%$ & $\approx 36$ \\ 
  & 8 & $97.16\%$ & $\approx 31$ \\ 
  \hline
  & 3 & $96.31\%$ & $\approx 31$ \\ 
  SDFL & 4 & $97.43\%$ & $\approx 34$ \\ 
  & 6 & $97.22\%$ & $\approx 35$ \\ 
  & 8 & $97.06\%$ & $\approx 40$ \\  
 \hline
 \end{tabular}
 \end{center}
\end{table}

\begin{figure*}[!ht]
 \centering
 \subfigure[\label{fig:fl4:accuracy} Accuracy]{
 \psset{xunit=1mm,yunit=1mm,runit=1mm}
 \scalebox{1.2}{
\begin{pspicture}(0,0)(31,27)
 \scriptsize
 \put(-10.5,0){\includegraphics[clip,trim=0mm 0mm 5mm 0mm, height=32mm]{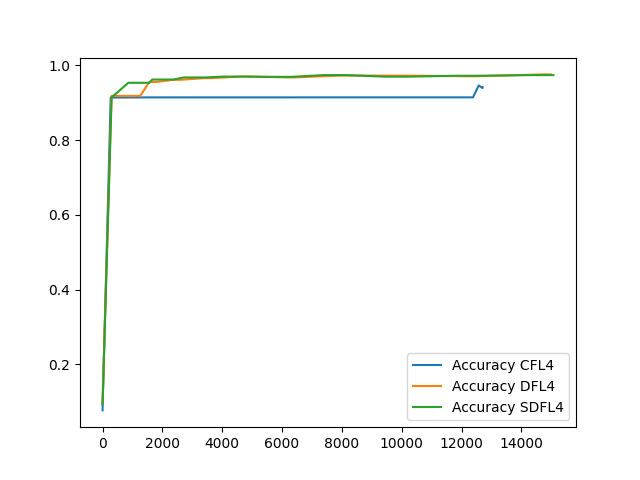}}
 \end{pspicture}
 }}
 \subfigure[\label{fig:fl4:loss} Loss]{
 \psset{xunit=1mm,yunit=1mm,runit=1mm}
 \scalebox{1.2}{
\begin{pspicture}(0,0)(31,27)
 \scriptsize
 \put(-5.5,0){\includegraphics[clip,trim=2mm 0mm 0mm 0mm, height=32mm]{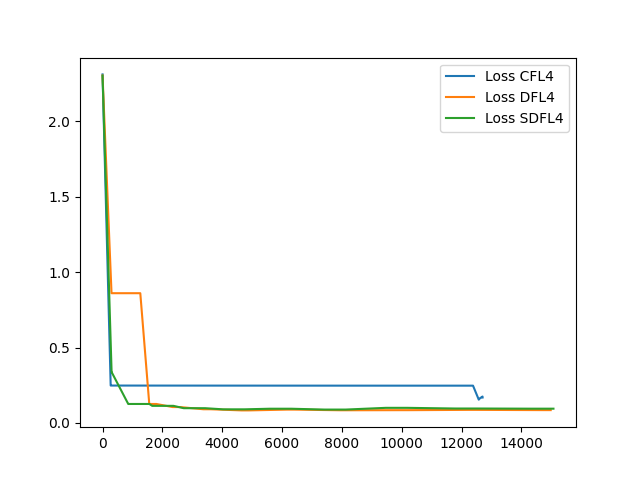}}
 \end{pspicture}
 }}
  \subfigure[\label{fig:fl4:bytes} Bytes exchanged]{
 \psset{xunit=1mm,yunit=1mm,runit=1mm}
 \scalebox{1.2}{
\begin{pspicture}(0,0)(31,27)
 \scriptsize
 \put(-2.5,0){\includegraphics[clip,trim=0mm 0mm 0mm 0mm, height=32mm]{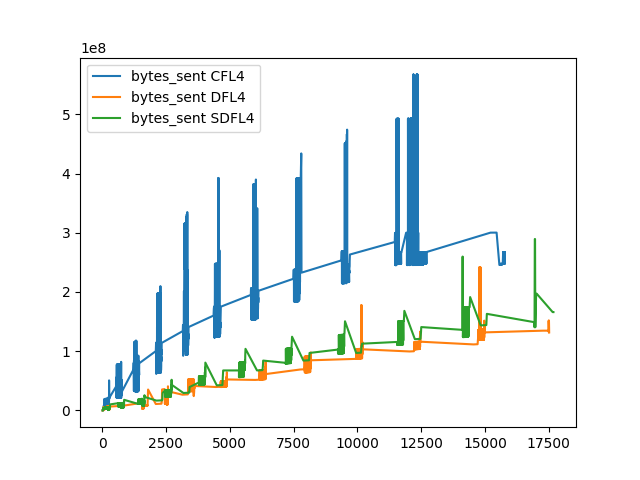}}
 \end{pspicture}
 }}
 \subfigure[\label{fig:fl4:cpu} CPU usage percentage]{
 \psset{xunit=1mm,yunit=1mm,runit=1mm}
 \scalebox{1.2}{
\begin{pspicture}(-5,0)(31,27)
 \scriptsize
 \put(-3.5,0){\includegraphics[clip,trim=0mm 0mm 0mm 0mm, height=32mm]{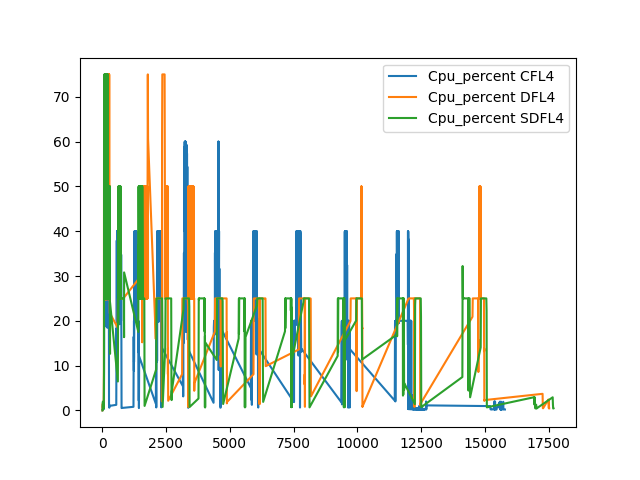}}
 \end{pspicture}
 }}
 \caption{Performance Evaluation Metrics with $|N|=4$ : (a) Accuracy  (b) Loss (c) Bytes exchanged in the federation network (d) CPU usage percentage. The DFL and the SDFL are the most performant architectures with an accuracy of 97.61\% and 97.43\% respectively. However, the SDFL is faster in term of convergence in comparison with the DFL with a loss of 0.08832. The SDFL is being close to the DFL in term of resource usage but the DFL still the less expensive architecture in term of communication. The CFL stays to the faster one in comparision with others and it is more stable but it is less performant than DFL and SDFL with an accuracy of 91.47\% and a loss of 0.154246.}\label{fig:fl4}
 \end{figure*}

\begin{figure*}[!ht]
 \centering
 \subfigure[\label{fig:fl6:accuracy} Accuracy]{
 \psset{xunit=1mm,yunit=1mm,runit=1mm}
 \scalebox{1.2}{
\begin{pspicture}(0,0)(31,27)
 \scriptsize
 \put(-10.5,0){\includegraphics[clip,trim=0mm 0mm 0mm 0mm, height=32mm]{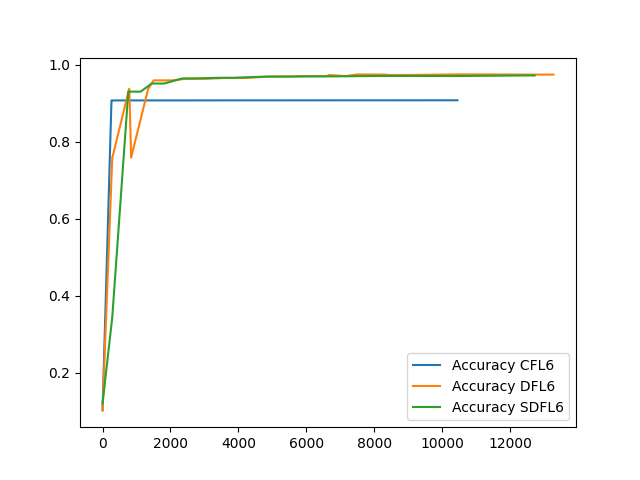}}
 \end{pspicture}
 }}
 \subfigure[\label{fig:fl6:loss} Loss]{
 \psset{xunit=1mm,yunit=1mm,runit=1mm}
 \scalebox{1.2}{
\begin{pspicture}(0,0)(31,30)
 \scriptsize
 \put(-5.5,0){\includegraphics[clip,trim=2mm 0mm 0mm 0mm, height=32mm]{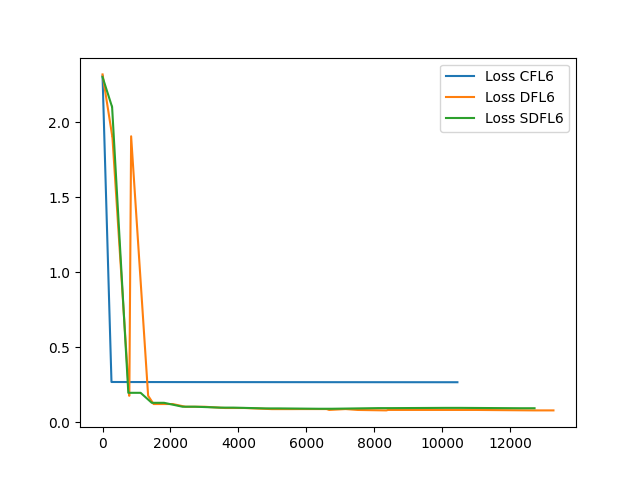}}
 \end{pspicture}
 }}
  \subfigure[\label{fig:fl6:bytes} Bytes exchanged]{
 \psset{xunit=1mm,yunit=1mm,runit=1mm}
 \scalebox{1.2}{
\begin{pspicture}(0,0)(31,27)
 \scriptsize
 \put(-2.5,0){\includegraphics[clip,trim=0mm 0mm 0mm 0mm, height=32mm]{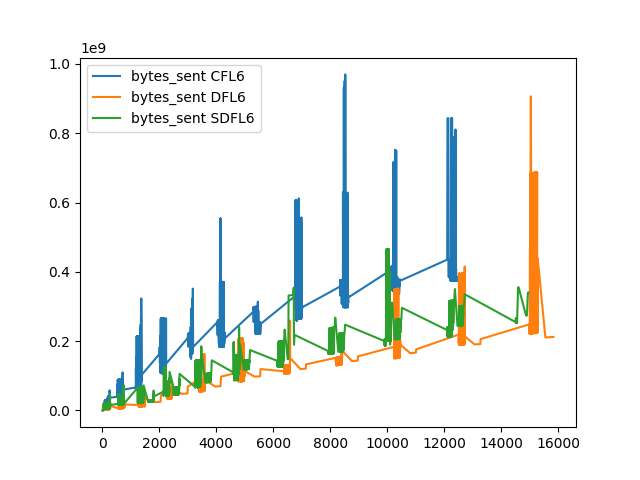}}
 \end{pspicture}
 }}
 \subfigure[\label{fig:fl6:cpu} CPU usage percentage]{
 \psset{xunit=1mm,yunit=1mm,runit=1mm}
 \scalebox{1.2}{
\begin{pspicture}(-5,0)(31,27)
 \scriptsize
 \put(-3.5,0){\includegraphics[clip,trim=0mm 0mm 0mm 0mm, height=32mm]{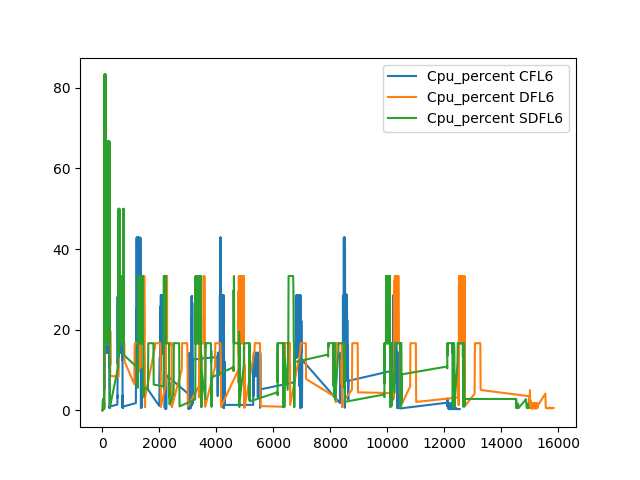}}
 \end{pspicture}
 }}
 \caption{Performance Evaluation Metrics with $|N|=6$ : (a) Accuracy  (b) Loss (c) Bytes exchanged in the federation network (d) CPU usage percentage. The DFL and the SDFL are the most efficient architectures with an accuracy of 97.5\% and 97.22\% respectively. However, the SDFL is faster in term of convergence in comparison with the , which is (the DFL) the less communication costing architecture, with a loss of   0.090313. The SDFL is being close to the DFL in term of resource usage. The CFL stays to the faster one in comparison with others and it is more stable but it is less performant than DFL and SDFL with an accuracy of 90.78\% and a loss of  0.266932.}\label{fig:fl6}
 \end{figure*}

\subsection{Results and analysis} \label{Sec:Results:res}
In the following, results obtained for the three FL architectures (CFL, DFL, SDFL) with respectively $3$, $4$, $6$, and $8$ participants in the FL aggregation process are commented. Let us also mention that all the results represent the average of all the participants' results collected from the experiments.

\figurename~\ref{fig:fl3} gives insight into the measured KPIs when using a network of $3$~participants in the FL process. It can be observed that the DFL architecture exhibits the highest performance, achieving an accuracy of 97.84\% \figurename~\ref{fig:fl3:accuracy}, a precision of 97.82\% \figurename~\ref{fig:fl3:precision}, a recall of 97.85\% \figurename~\ref{fig:fl3:recall} with the lowest loss value of 0.0729 \figurename~\ref{fig:fl3:loss} (\tablename~\ref{table:1} shows the test results of some evaluation metrics for the implemented scenarios). Meanwhile, the semi decentralized federated learning (SDFL) architecture closely approaches the performance of DFL with an accuracy of 96.31\% \figurename~\ref{fig:fl3:accuracy}, a precision of 96.28\%, a recall of 96.31\% and a loss value of 0.1334, requiring less time (about $6~min$ according to \tablename~\ref{table:1}). Additionally, the CFL architecture demonstrates faster convergence to performance levels compared to the others (about half of the time according to \tablename~\ref{table:1}) with less efficient performances (accuracy of 94.45\% \figurename~\ref{fig:fl3:accuracy}), showcasing greater stability. This outcome aligns with expectations, as CFL is inherently more resource-intensive (according to \figurename~\ref{fig:fl3:bytes} and \figurename~\ref{fig:fl3:cpu}) due to its reliance on a central server entity for model aggregation, while DFL stands out as more lightweight, eliminating the need for an additional aggregation entity (\figurename~\ref{fig:fl3:bytes}).

\begin{figure*}[!ht]
 \centering
 \subfigure[\label{fig:fl8:accuracy} Accuracy]{
 \psset{xunit=1mm,yunit=1mm,runit=1mm}
 \scalebox{1.2}{
\begin{pspicture}(0,0)(31,27)
 \scriptsize
 \put(-10.5,0){\includegraphics[clip,trim=0mm 0mm 5mm 0mm, height=32mm]{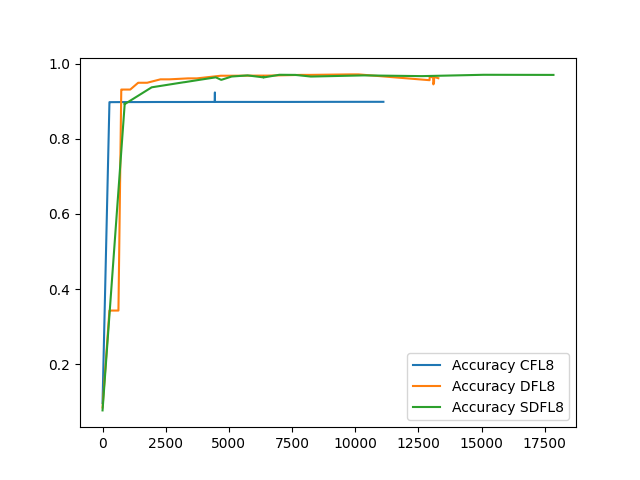}}
 \end{pspicture}
 }}
 \subfigure[\label{fig:fl8:loss} Loss]{
 \psset{xunit=1mm,yunit=1mm,runit=1mm}
 \scalebox{1.2}{
\begin{pspicture}(0,0)(31,30)
 \scriptsize
 \put(-5.5,0){\includegraphics[clip,trim=2mm 0mm 0mm 0mm, height=32mm]{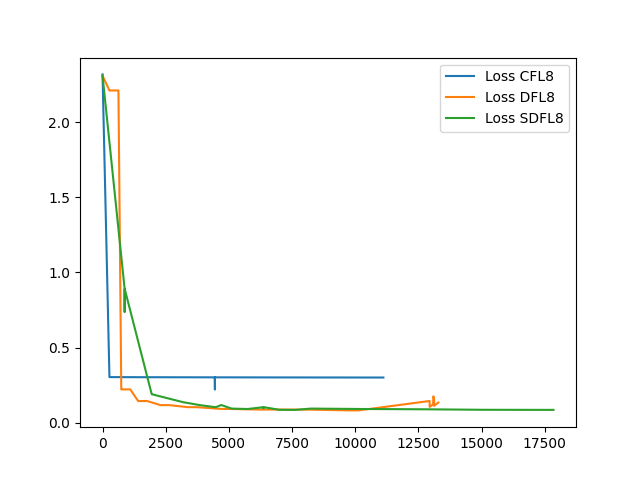}}
 \end{pspicture}
 }}
  \subfigure[\label{fig:fl8:bytes} Bytes exchanged]{
 \psset{xunit=1mm,yunit=1mm,runit=1mm}
 \scalebox{1.2}{
\begin{pspicture}(0,0)(31,27)
 \scriptsize
 \put(-2.5,0){\includegraphics[clip,trim=0mm 0mm 0mm 0mm, height=32mm]{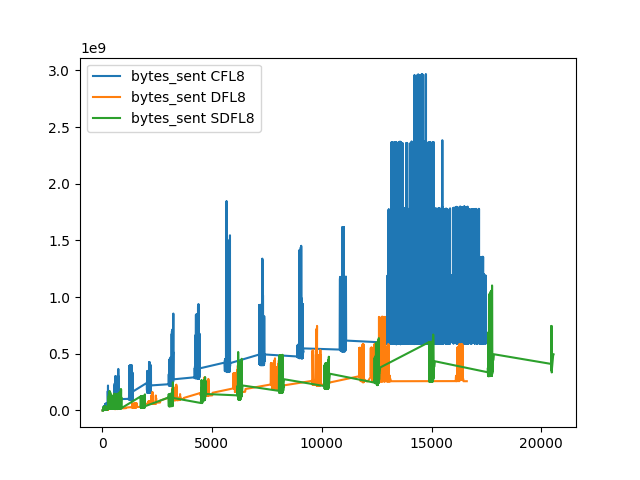}}
 \end{pspicture}
 }}
 \subfigure[\label{fig:fl8:cpu} CPU usage percentage]{
 \psset{xunit=1mm,yunit=1mm,runit=1mm}
 \scalebox{1.2}{
\begin{pspicture}(-5,0)(31,27)
 \scriptsize
 \put(-3.5,0){\includegraphics[clip,trim=0mm 0mm 0mm 0mm, height=32mm]{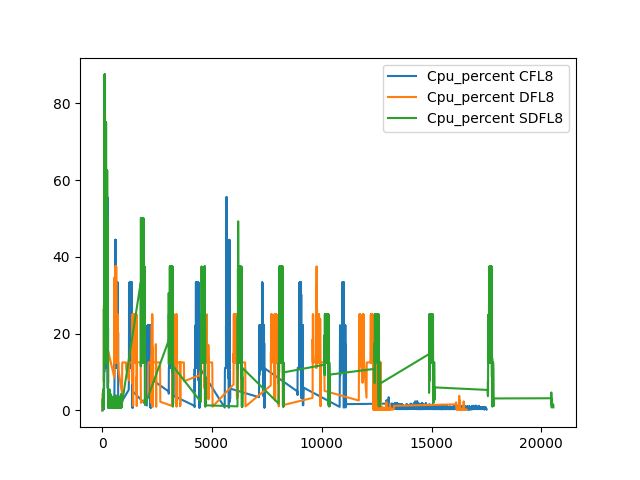}}
 \end{pspicture}
 }}
 \caption{Performance Evaluation Metrics with $|N|=8$ : (a) Accuracy  (b) Loss (c) Bytes exchanged in the federation network (d) CPU usage percentage. The DFL and the SDFL still the most performant architectures with an accuracy more than 97\% . However, the SDFL is faster in term of convergence in comparison with the DFL one. The SDFL is being really close to the DFL in term of resource usage. The CPU usage is being almost the same for all scenarios: CFL, DFL and SDFL, but still the CFL is more expensive when it comes to the communication cost with less model performance (accuracy of 89.83\%).}\label{fig:fl8}
 \end{figure*}

As evidenced in \figurename~\ref{fig:fl4}, \ref{fig:fl6}, and \ref{fig:fl8}, when scaling the number of participants respectively to $4$, $6$ and $8$ in the learning process, both DFL and SDFL architectures exhibit superior performance, boasting precision exceeding 97\% according to \figurename~\ref{fig:fl4:accuracy}, \figurename~\ref{fig:fl6:accuracy}, \figurename~\ref{fig:fl8:accuracy} and \tablename~\ref{table:1} (with a marginal difference of approximately 0.2\% between them) with the minimum loss values as shown in \figurename~\ref{fig:fl4:loss}, \ref{fig:fl6:loss}, and \ref{fig:fl8:loss} when comparing them to CFL. Let us note that, concerning the model's performance, we included only the accuracy results involving 4, 6, and 8 participants due to the page constraints. One may also observe that SDFL converges faster than DFL when the participant count is 4 and 6 (see \figurename~\ref{fig:fl4:accuracy} and \ref{fig:fl6:accuracy}). However, when scaled to 8 participants, DFL demonstrates faster convergence compared to SDFL \figurename~\ref{fig:fl8:accuracy} and \tablename~\ref{table:1}. This observation was somehow expected, given that DFL lacks of a global aggregation process, which typically contributes to its accelerated convergence. Despite this difference in convergence rates, the communication costs between DFL and SDFL remain closely aligned throughout the scaling process according to \figurename~\ref{fig:fl4:bytes}, \ref{fig:fl6:bytes} and \ref{fig:fl8:bytes}. Furthermore, despite the fact that CFL is characterized as resource-intensive, it maintains its position as the swiftest in achieving performance convergence, showcasing greater stability.

\begin{figure*}[!ht]
 \centering
 \subfigure[\label{fig:fl:accuracy_cfl} Accuracy CFL]{
 \psset{xunit=1mm,yunit=1mm,runit=1mm}
 \scalebox{1.2}{
\begin{pspicture}(0,0)(27,27)
 \scriptsize
 \put(-5.5,0){\includegraphics[clip,trim=0mm 0mm 2mm 0mm, height=30mm]{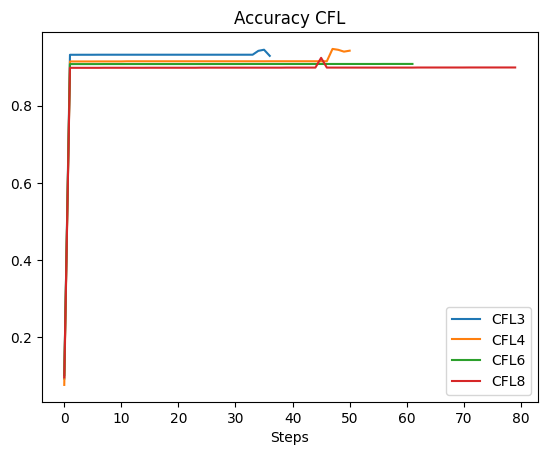}}
 \end{pspicture}
 }}
 \subfigure[\label{fig:fl:accuracy_dfl} Accuracy DFL]{
 \psset{xunit=1mm,yunit=1mm,runit=1mm}
 \scalebox{1.2}{
\begin{pspicture}(0,0)(31,30)
 \scriptsize
 \put(-.5,0){\includegraphics[clip,trim=2mm 0mm 0mm 0mm, height=30mm]{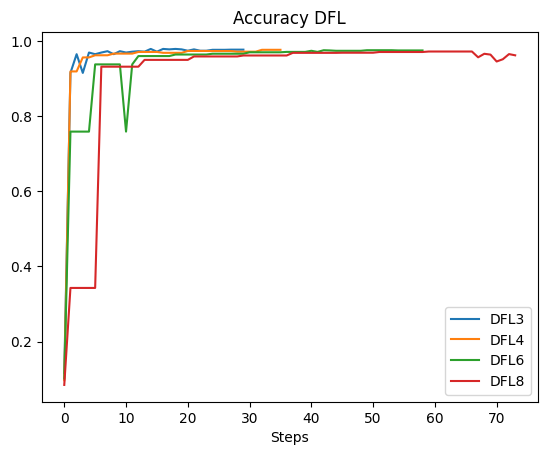}}
 \end{pspicture}
 }}
  \subfigure[\label{fig:fl:accuracy_sdfl} Accuracy SDFL]{
 \psset{xunit=1mm,yunit=1mm,runit=1mm}
 \scalebox{1.2}{
\begin{pspicture}(0,0)(27,27)
 \scriptsize
 \put(-.5,0){\includegraphics[clip,trim=0mm 0mm 0mm 0mm, height=30mm]{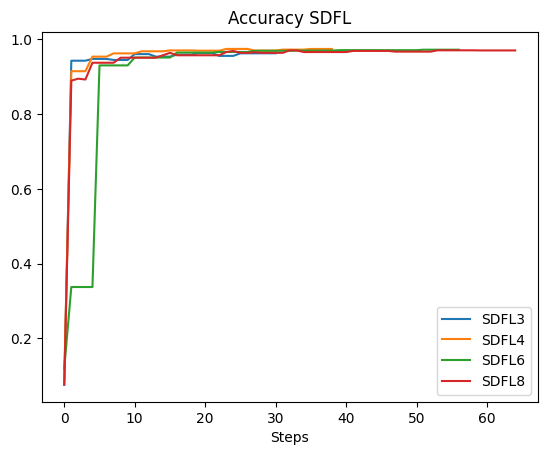}}
 \end{pspicture}
 }}
 \caption{Accuracy average in all scenarios: (a) CFL  (b) DFL (c) SDFL. In CFL, increasing the number of participants in the network leads to a decrease in model's performance. This occurs because each participant has less data samples. In contrast, the performance of the model DFL and SDFL remains practicaly the same. However, the convergence time of the model is prolonged in DFL and SDFL settings.}\label{fig:fl}
 \end{figure*}

In centralized (CFL) setups, a notable trend can be observed: as the number of participants increases, there is often a decline in model performance. This phenomenon arises due to the diminishing number of samples assigned to each participant. With a broader participant pool, the dataset gets divided into smaller subsets, constraining the volume of data accessible for training. Consequently, this decrease in sample size impedes the model's capacity to generalize effectively, leading to diminished performance.

In contrast to DFL and SDFL architectures, the number of clients participating in the federation does not significantly impact the performance of the model. This resilience stems from the distributed nature of these architectures, where each participant retains a local dataset and computes model updates independently. Thus, the model's performance remains stable regardless of the number of participants. Nevertheless, it is important to highlight that although DFL and SDFL demonstrate consistent performance across different participant counts, they usually demand more time to converge (cf., \tablename~\ref{table:1}). This elongated convergence period is ascribed to the decentralized nature of these architectures, which necessitates coordination and communication among participants to aggregate model updates. Despite the extended convergence time, DFL and SDFL confer the benefit of resilience to fluctuations in participant count, thereby ensuring stable model performance over time.

\section{Conclusion}\label{Sec:conclu}
This paper addresses the lack of experimental studies between the three types of Federated Learning (FL) architectures, Centralized FL (CFL), Decentralized FL (DF) and Semi-Decentralized FL (SDFL). Although the choice of the type of architecture is highly dependent on the application needs and constraints, simulating the different architectures to understand their limitations (and benefits) in terms of performance is important. This paper is an attempt to bring into light possible trade-offs between distinct performance indicators (aka KPIs). Our experiments show that DFL emerges as a promising strategy, offering enhanced performance while maintaining data privacy. Conversely, in larger networks, SDFL presents itself as a cost-effective solution, delivering comparable performance with reduced resource requirements. However, when speed is of paramount importance, CFL remains the optimal choice for achieving fast results. In essence, DFL, SDFL and CFL exhibit strengths and weaknesses. The best option depends on factors like the nature of the data, the number of nodes/clients, privacy considerations, and infrastructure availability requirements or constraints. 

Looking ahead, our future research aims to cover a wider range of KPIs, and explore further trade-offs among KPIs in diverse FL applications. This will require to extend the code of the FedStellar platform. Furthermore, we plan to develop and release a decision support tool to guide application and network managers in selecting the most appropriate FL architecture (and possibly platform) depending on their application needs, requirements, and constraints.

\end{document}